# Delving Deep into Liver Focal Lesion Detection: A Preliminary Study

Jiechao Ma[1,3#] Yingqian Chen[2#], Yu Chen[1], Fengkai Wan[3], Sumin Xue[3], Ziping Li[2*], Shiting Feng[2*]

[1] Sun Yat-sen University, Guangzhou, China

[2] The First Affiliated Hospital, Sun Yat-sen University, Guangzhou, China

[3] Infervision Inc., Beijing, China

Jiechao Ma#,[1,3#] Yingqian Chen[#] (Co-first author, equal contribution)

Correspondence should be addressed to

Shiting Feng: fst1977@163.com

Ziping Li: liziping163@163.com

## Abstract

Hepatocellular carcinoma (HCC) is the second most frequent cause of malignancy-related death and is one of the diseases with the highest incidence in the world. Because the liver is the only organ in the human body that is supplied by two major vessels: the hepatic artery and the portal vein, various types of malignant tumors can spread from other organs to the liver. And due to the liver masses' heterogeneous and diffusive shape, the tumor lesions are very difficult to be recognized, thus automatic lesion detection is necessary for the doctors with huge workloads. To assist doctors, this work uses the existing large-scale annotation medical image data to delve deep into liver lesion detection from multiple directions. To solve technical difficulties, such as the image-recognition task, traditional deep learning with convolution neural networks (CNNs) has been widely applied in recent years. However, this kind of neural network, such as Faster Regions with CNN features (R-CNN), cannot leverage the spatial information because it is applied in natural images (2D) rather than medical images (3D), such as computed tomography (CT) images. To address this issue, we propose a novel algorithm that is appropriate for liver CT imaging. Furthermore, according to radiologists' experience in clinical diagnosis and the characteristics of CT images of liver cancer, a liver cancer-detection framework with CNN, including image processing, feature extraction, region proposal, image registration, and classification recognition, was proposed to facilitate the effective detection of liver lesions.

## Introduction

At present, China's medical and health data are growing explosively, and the healthcare structure has changed from a central, large-scale system to a universal, individual-centric one. The medical and health service industry faces unprecedented challenges [1]. Liver lesion detection is one of the key issues in medical imaging and has become a hot topic in this field [2–4]. With the rapid development of computer vision and artificial intelligence, along with medical imaging and computer-assisted intervention, researchers have found that the use of computer-assisted diagnostic systems in medicine can largely reduce their workload and assist them in diagnosis [5,6]; therefore, medical image-detection problems, like liver lesion detection, are receiving considerable attention.

Until now, in the medical field, research on liver tumors primarily remains at the stage of manual extraction of features [7,8], and most studies only consider a single mode. Although there have been many good manual feature extractors with invariance and separability, like scale-invariant feature transform (SIFT) [9], these manual features do not have good universality and there is a need for human interactive methods to obtain an acceptable result. Therefore, a deep-learning method based on the automatic extraction of feature patterns is proposed [10–14].

Recently, deep-learning techniques have been applied to a wide variety of medical image problems. The traditional neural network, Artificial Neural Networks (ANNs), is a machine-learning technique [15] inspired by the study of brain modelling, a mathematical model for distributed parallel information processing. While the deep-learning algorithm is a major breakthrough in the field of ANNs. By combining different low-level features to form more abstract high-dimensional features, the abstract distributed characteristics of the data can be discovered [16]. The basic flow of the deep-learning model algorithm is usually follows the machine learning methods to solve the problem. The general idea is a pre-process of the data obtained by the sensor, because the original input is difficult to express in the deep-learning framework.

In general, deep learning is a simple way to normalize the pixels of an image so that they are all in the same distribution. For the pre-processed images, however, the extracted images can represent the features of the images themselves, but the quality of the extracted features significantly determines the accuracy of the task. In the final step, it was inferred that the category of the objects in the image is the core part of deep learning, and most of the current

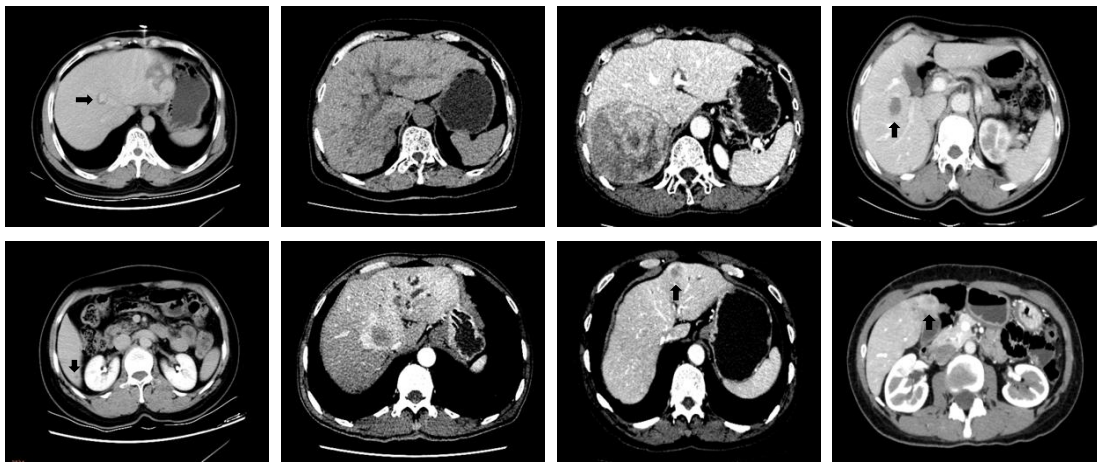

work focuses on this step.

Figure 1: Different Challenges for Liver Lesion Detection from CT Data

In terms of clinical diagnosis, due to certain characteristics of the liver, the liver lesion-detection task is still a great challenge [17,18].

First, because of the low contrast of liver lesions in computed tomography (CT), some lesions are difficult for doctors to find (shown in the first column of Figure 1). In order to solve this problem, contrast agents are needed. Images after enhancement are employed to increase the discrimination of liver lesions.

Second, the boundaries of the liver lesion are not clear (shown in the second column of Figure 1), which makes it difficult to find the lesion accurately with traditional detection algorithms. In addition, the resolution of the image data is not uniform, which is also one of the important factors that affects the accuracy of detection algorithms. Furthermore, the difference on the z-axis of the CT image is relatively large (the voxel spacing ranges from 0.45 mm to 6 mm), which is more difficult to automatically detect with the algorithm;

Third, due to the heterogeneous and diffusive shapes, and the different sizes of the same type of mass (shown in the third column of Figure 1), the tumor lesions are very difficult to recognize.

Fourth, there is interference between different lesions of the liver mass. The variety of the abdominal organs is a major obstacle for the development of automatic detection techniques for liver tumors. The liver masses can be divided into primary (i.e., originating in the liver) or secondary (i.e., spreading to the liver) lesions, including malignant lesions, such as hepatocellular carcinoma, and benign lesions, such as hemangiomas and cysts, which demonstrate morphological similarity in the CT images (shown in the fourth column of Figure 1).

To tackle these above difficulties, many researchers used typical detection methods in deep learning, such as the Faster Regions with convolutional neural network (CNN) features (R-CNN) algorithm [19,20].

Although Faster R-CNN can solve the above problems well, due to the increasing number of layers in the neural network, many of the low-level features are discarded, resulting in a low detection accuracy. On the other hand, for the diagnosis of liver lesions, it is usually performed in combination with multiple phases (such as the arterial and portal phases) of CT imaging, but Faster R-CNN is not capable of this.

To address the above problems, we proposed a novel detection framework, which can combine several multiple multi-scale feature descriptors with different resolutions to fuse different depths of information and propose a new algorithm to discover the relationships between the various modalities.

In summary, the aim of this study is to find a system that can automatically identify the features of the liver lesion to assist the doctor in diagnosis and provide theoretical and practical support for the treatment of cancer and improvement of people's health in practical applications, thereby making efforts for the optimization of health and medical services. Key technical contributions are as follows:

1. Seeking a better feature-extraction network to better fit the data by increasing the depth of the network to obtain a better system.

2. In the region-proposal phase of the framework, we improve the traditional deep-learning detection method by using multiple scales of features to improve the accuracy of the framework.

3. The proposed framework divides the 3D image into multiple slices and uses multiple methods to combine different slices to make it as 2.5D, which not only enables full use of the effects of spatial data, but also addresses the limitation of computing capacity.

4. Using the information of multi-phase CT images of the liver, we propose a novel algorithm of deep learning to register and fuse CT images at different phases to explore the interrelationship between different modalities.

**Materials and Methods**

The detection task based on deep neural networks is a complex problem that often exists in medical image processing [21–23]. In this chapter, we first define the overall outline of the liver lesion-detection tasks and introduce the data sets used in the experiments. We then describe in detail the detection framework constructed by the convolutional division network: we used a different number of layers of CNN to extract features, and then we used the algorithm on the extracted features to propose regions of interest (ROIs) of different scales. For these obtained ROIs, we used a neural network to detect and classify, offering the position and category of the lesion in the image. Finally, due to the specificity of the CT data of the liver, we combined a variety of modal data and performed experiments on the dataset based on this framework.

*Outline*

For the detection of liver lesions using neural networks, the model framework proposed in this paper is shown in Figure 2. For every neural network, the framework consists of a Training phase and a Testing phase. In the training phase, the obtained CT data was enhanced by using some methods called data augmentation (e.g., scaling) [24,25]. The enhanced data, called input data, were then entered into the neural network framework to obtain a trained framework. In the Feature Extraction stage, we tried to find a better feature extraction network by testing different numbers of layers of CNNs. In order to address the deficiencies in traditional neural-network recognition algorithms that cannot fully utilize the 3D spatial information, we used a 2.5D-like method to extract spatial information between different slices. In the Region Proposal stage, we used a pyramid structure to extract the proposals in order to capture different sizes of lesions. We call such a proposal the ROI. In this stage, we designed a classifier based on texture features to divide the ROIs into normal and abnormal liver lesions. Next, at the stage of Classification detection, we used abstract features to divide abnormal liver lesions into hepatocellular carcinomas (HCCs), liver cysts, and hemangiomas. For this process, we conducted multiple iterations during the Training phase to obtain a better framework model. Finally, in the Testing phase, we used another batch of CT imaging to evaluate the framework based on the results.

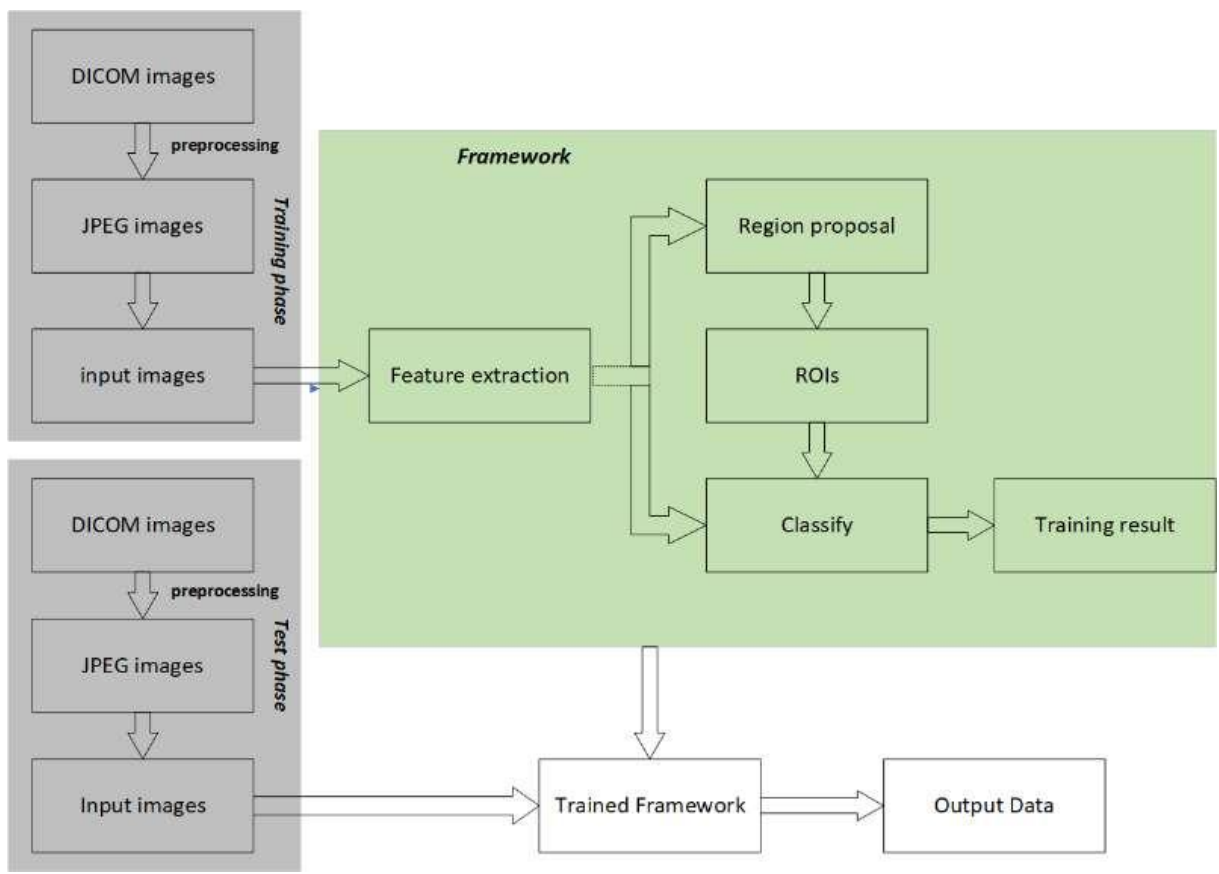

Figure 2: The illustration of the pipeline for liver lesion detection

*CT Imaging*

The clinical retrospective study used sets of CT images of liver masses over three phases (non-contrast-agent enhanced, arterial, and delayed) [26]. In general, a mass that only exists in a single period cannot be used to determine the type of lesion. Therefore, it is necessary to confirm the diagnosis by comparing the contrast changes between the periods by contrast injection; The time settings of the high-pressure syringe were (from the injection time, with an injection speed of approximately 3.0 ml/s) 25 s for the arterial phase, 50–55 s for the portal phase, and 115 s for the venous phase. Each modality contained a number of two-dimensional tomographic images, which can be simply viewed as a three-dimensional model. Both the training set and the testing set are CT abdominal scans that were digitally imaged using a Digital Diagnost Release 1.3 digital camera without deep exposure, and the exposure conditions are 120 kVp, 3 mAs; the camera used was Toshiba, GE healthcare. The scan was performed using 128iCT. The patient was examined in a supine position, with arms raised and breath held, and was scanned after deep breathing. The scanning range is from the tip of the liver to the end of the liver. The canning parameters were 120 kVp, 80 mA, with the reconstruction algorithm of a full iterative model reconstruction, and 1-mm reconstruction thickness and 1-mm layer spacing. All images were coronal reconstructed and compared with the postero-anterior (P-A) position of the lesion.

*Deep Learning with CNN*

The system environment used in this study was Ubuntu 14.04. The deep-learning and computing unit comprises four GEFORCE GTX 1080 GPUs with 8-GB GPU memory. The central processing unit is the Intel i7-7700. The high-speed memory is 32-GB RAM. The image storage unit is SAMSUNG 128GB. The storage security unit comprises two SEAGATE 3-TB drives; and the deep-learning framework used in this article is Mxnet 1.0.1 (http://mxnet.incubator.apache.org/) [27], which is based on C++/CUDA/Python. In the training phase of the network, all pre-processed data were input to a neural network for training, and then the model was updated through learning iterations using the loss value calculated by the back-propagation method; thus, two outputs of the network could be obtained: lesion locations and lesion categories. In addition, this work also used a series of methods to increase the accuracy of the framework (transfer learning [28], Dropout [29,30], Batch Normalization [31], etc.). Transfer Learning uses the knowledge learned from one environment to solve tasks in a new one. In general, the key issue is to use existing knowledge to assist in learning new knowledge as early as possible. The core issue is to find the similarities between existing knowledge and new knowledge. Dropout can be considered a method of using bagging technology for a highly integrated neural network. The main idea of the algorithm is to randomly disable some nodes in the network during the training of the model. In the next iteration, these nodes are activated again. For training, the previous two steps are repeated and several new random subnets are formed, which can finally enhance the training result especially when the training data is limited. Batch Normalization is a method that can almost reparametrize deep networks, which can improve the coherence between the update of difference layers in the neural network.

*Image Processing*

In this work, we used the RadiAnt DICOM Viewer 4.2.0 software (https://www.radiantviewer.com/dicom-viewer-manual/v/4.2.0/) to display the DICOM image. The DICOM image was converted into an 8-bit Joint Photographic Experts Group (JPEG) format with an image size of 512×512 pixels. The programming language used for the image data was Python 2.7.6 (https://www.python.org/), and the image matrix processing library was NUMPY 1.13.3 (http://www.numpy .org/). For all training and test image data, we used a pixel-level interpolation method to convert the image resolution to 800×800 pixels. For the DICOM data-specific properties (window width, window level), we made appropriate adjustments, such as converting to the window width 80 HU, window level 150 HU, and then converting it to the RGB value of the JPEG format. In order to solve the problem of overfitting due to the lack of a large amount of data, many data-augmentation methods were used. For data labelling, all the lesion labels were jointly discussed by two experienced radiologists, and all the labels were completed according to the abdominal diagnostic report. For the data labelling tool, customized modification and development have been made for medical images.

*Image Data: Input Data*

In medical images, tumors have an unpredictable shape and can be found anywhere in the liver. Moreover, even two tumors of the same type may look very different. In general, depending on the incidence, tumors can be divided into the following categories: classic hepatocellular carcinomas (HCCs) and other malignant liver tumors, high incidence of benign masses (hemangiomas and liver cysts), and other benign masses. This study used these three lesions with the highest incidence rate to carry out experiments to verify the validity of the proposed framework. For each patient, there were three corresponding phase CT images (non-contrast-agent enhanced, arterial).

Table 1: Diagnostic Details for Each Category of Liver Mass

| Training Set | No. Patients | No. Lesions | Test Set | No. Patients | No. Lesions |
|---|---|---|---|---|---|

| Liver Cyst | 178 | 910 | - | 20 | 98 |
| Hemangiomas | 353 | 495 | - | 20 | 38 |
| HCCs | 90 | 173 | - | 20 | 30 |

As seen in Table 1, we used 621 patients to train the model, and the training set included 910 cyst lesions, 495 hemangiomas, and 173 HCCs. We then used 60 other patients to test the model.

*Feature Extraction*

In the feature extraction phase, we used several traditional deep-learning feature extraction networks that are well represented in the field of computer vision [34–37]. Because the number of layers in the deep-learning network has a significant influence on the final classification and recognition results, the normal method is to design the network as deep as possible. In order to verify this idea, we used neural networks with 50 layers and 101 layers to compare the results of the framework.

For traditional neural-network feature extraction, the spatial information between different slices cannot be captured effectively. In order to better find the spatial information between different slices, we fused the adjacent 9 slices of the CT data, forming an approximately 2.5D structure, and then we used this 2.5D-like data for feature extraction.

*Region Proposal*

As the neural network becomes deeper and deeper, its low-level information is gradually erased [32,33] and the feature map image tends to become more abstract. Therefore, the correlation information between low-level pixels in the deep structure of the network is often missed [38–41]. The best way to solve this problem is to add low-dimensional pixel-level information into high-dimensional space information. Therefore, this study combined the high-dimensional spatial information and the low-dimensional information space for the proposed structure during the region-proposal stage of the network.

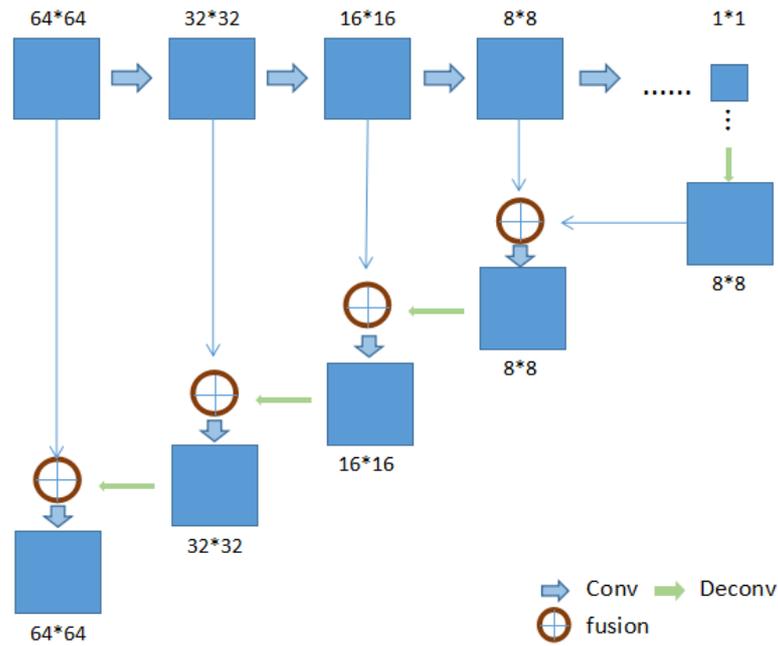

Figure 3: Illustration of the fusion method

The fusion method, as shown in Figure 3, shows that for deeper feature maps (for example, $1\times1$-feature maps) at the network layer, the information is more meaningful, such as a cyst. However, the shallow feature map of the layer is more local (for example, $64\times64$-feature map) and may be a vertical or horizontal texture of the cyst. Thus, deep information is often formed by shallow information.

*Classification Recognition*

In the training phase, each ROI generated in the region proposal stage has a corresponding label that is created by the radiologists. For each ROI, the classifier must indicate to which category the ROI belongs and where the corresponding position of the ROI is in the original image. However, we took the ROI with the highest probability as the output of the model (as shown later in Figure 5). During the testing phase, we used the doctor's annotation as an evaluation of the performance of the algorithm.

*Image Registration*

However, because there are four common types of contrast (non-contrast-agent enhanced, arterial, and delayed) for liver CT images, each modality contains a number of two-dimensional tomographic images and each modality has a non-negligible role in the diagnosis of lesions. We proposed an unprecedented deep-learning image registration algorithm. In order to find the association information between two modalities, we defined a relational equation.

$$L = \frac{1}{C(x)} \sum f(x,y) g(x) \qquad\qquad (1)$$

In Equation 1, we defined x and y as two different modalities, where f(x) and g(x) are two different relational functions. $L$ can be defined as the relationship between two modalities. We added this kind of relationship in deep neural networks to mine the association between different modalities. From another perspective, this is the registration of data from different modalities.

**Results and Discussion**

In this chapter, we solve some of the problems that exist in traditional neural networks in the liver lesion-detection task, then analyzed the influence of several improvements we have proposed in the framework. Through comprehensive analysis of the influence of various components on the model framework, we obtained a better deep neural-network model. Finally, according to the proposed model, a large number of experiments were conducted on the test set to demonstrate the validity of the framework.

1. *Effectiveness of the depth of CNN*: In order to verify the influence of the number of deep neural networks on the model, we proposed two networks with large differences in the number of layers. The detailed structure of the networks are described in Table 2 and Table 3.

Table 2: Detailed Architecture of 50-Layer Proposed Network

|  | Layer | Feature size |
|---|---|---|
| Conv1 | 7×7, 64, stride 2, 2, 2 | 16×224×224 |
| Pool1 | 3×3×3 max, stride 2, 2, 2 | 8×112×112 |
| Block1 | $\begin{bmatrix} 1\times1 & 64 \\ 3\times3 & 64 \\ 1\times1 & 256 \end{bmatrix} \times 3$ | 8×112×112 |
| Pool2 | 3×1×1 max, stride 2,1,1 | 4×112×112 |
| Block2 | $\begin{bmatrix} 1 \times 1 & 128 \\ 3 \times 3 & 128 \\ 1 \times 1 & 512 \end{bmatrix} \times 4$ | 4×56×56 |
| Block3 | $\begin{bmatrix} 1 \times 1 & 256 \\ 3 \times 3 & 256 \\ 1 \times 1 & 1024 \end{bmatrix} \times 6$ | 4×28×28 |
| Block4 | $\begin{bmatrix} 1 \times 1 & 512 \\ 3 \times 3 & 512 \\ 1 \times 1 & 2048 \end{bmatrix} \times 3$ | 4×14×14 |
| Concat | - | 4×224×224 |

It can be seen from Table 2 that the model structure is composed of the convolution layer and the pooling layer in a deep neural network. In the block layer, we stacked a number of convolution and pooling layers to form this block, and there are three layers in each block. Then, these blocks were repeated at this stage. For example, the block was repeated three times in the block$_1$ phase, so there were 9 layers in this block$_1$.

Table 3: Detailed Architecture of 101-Layer Proposed Network

| | Layer | Feature size |
|---|---|---|
| Conv1 | 7×7, 64, stride 2, 2, 2 | 16×224×224 |
| Pool1 | 3×3×3 max, stride 2, 2, 2 | 8×112×112 |
| Block1 | $\begin{bmatrix} 1 \times 1 & 64 \\ 3 \times 3 & 64 \\ 1 \times 1 & 256 \end{bmatrix} \times 6$ | 8×112×112 |
| Pool2 | 3×1×1 max, stride 2,1,1 | 4×112×112 |
| Block2 | $\begin{bmatrix} 1 \times 1 & 128 \\ 3 \times 3 & 128 \\ 1 \times 1 & 512 \end{bmatrix} \times 8$ | 4×56×56 |
| Block3 | $\begin{bmatrix} 1 \times 1 & 256 \\ 3 \times 3 & 256 \\ 1 \times 1 & 1024 \end{bmatrix} \times 12$ | 4×28×28 |
| Block4 | $\begin{bmatrix} 1 \times 1 & 512 \\ 3 \times 3 & 512 \\ 1 \times 1 & 2048 \end{bmatrix} \times 6$ | 4×14×14 |
| Concat | - | 4×224×224 |

From Table 3, we constructed a 101-layer convolutional neural network with a structure similar to that of the above-mentioned 50-layer neural network. However, the difference is that we added a transition layer to each block to maintain the feature size. In the experiment, we compared the 50-layer convolutional neural network and the 101-layer convolutional neural network to the model framework. It can be found from Table 4 that under the same conditions, the accuracy of the 101-layer network is significantly improved compared to that of the 50-layer network.

Table 4: The predictive accuracy of different frameworks

|  | Liver Cyst (%) | Hemangiomas (%) | HCCs (%) |
|---|---|---|---|
| R-50 | 69.23 | 64.29 | 70.96 |
| R-101 | 78.26 | 72.91 | 76.93 |
| R-50-2.5D | 72.71 | 66.23 | 71.56 |
| R-101 region fusion | **83.33** | 78.35 | 76.15 |
| R-50 multi-modal | 72.47 | 68.43 | 72.14 |
| R-101 multi-modal | 82.76 | **78.98** | **77.78** |

2. *Effectiveness of the 2.5D-like feature*: Because medical images are mostly based on 3D datasets, the complexity of their data can be imagined. Therefore, modern computing power is a major bottleneck in medical images. Many scholars have sought solutions to this issue. This includes the operation of dividing the 3D medical image into different patches. However, for such a method, the effect of promotion is limited because the input image does not include some global information and spatial information. The framework proposed in this paper divides the 3D image into multiple slices and combines different adjacent slices. It not only fully utilizes the effect of medical image spatial information, but also solves the limitation of the computing capacity. It can be found from Table 4 that the 2.5D-like structure can improve the model to a certain extent. In addition, we can see that cysts and hemangiomas increase more than HCCs in terms of accuracy because cyst and hemangiomas are relatively small, and therefore the use of the context information from several slices can cover the entire lesion.

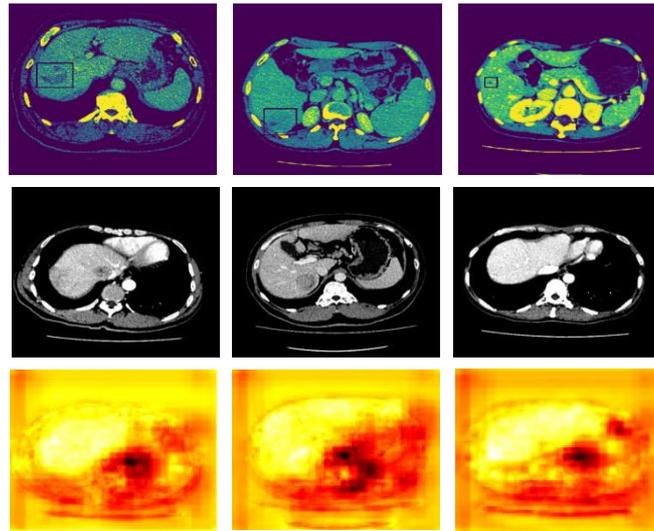

Figure 4: Illustration of the different phase features

3. *Effectiveness of the region fusion*: The experiment exploited the characteristics of a multiscale-based deep network approach that enables the combination of high-dimensional and low-dimensional information to become more sensitive to small objects. It can be found from Table 4 that the model using the region fusion method is more sensitive to smaller objects. This makes the model greatly improved for use detecting small objects. This can also explain the model's great improvement in the accuracy rate of cysts and hemangiomas in contrast to that of HCCs.

4. *Effectiveness of the multi-modalities*: For medical data, multimodal imaging data places higher demands on neural-network algorithms. The experiments used a neural network for different modalities to learn the association information between them, thereby correlating the image data between different modalities. It can be found from Table 4 by comparing the 5[th] and 6[th] row with the 3[rd] and 4[th] ones: : The results show that the proposed multimodality method can achieve better results in hemangiomas and HCC. Because the feature of cysts in each modality does not change much, the relationship network cannot effectively express the change in cysts. This causes the method to have little effect on the accuracy of cysts and also confirms that the relationship network proposed in this paper plays an important role in the detection of lesions that can change over time. The relationship feature maps are shown in Figure 4.

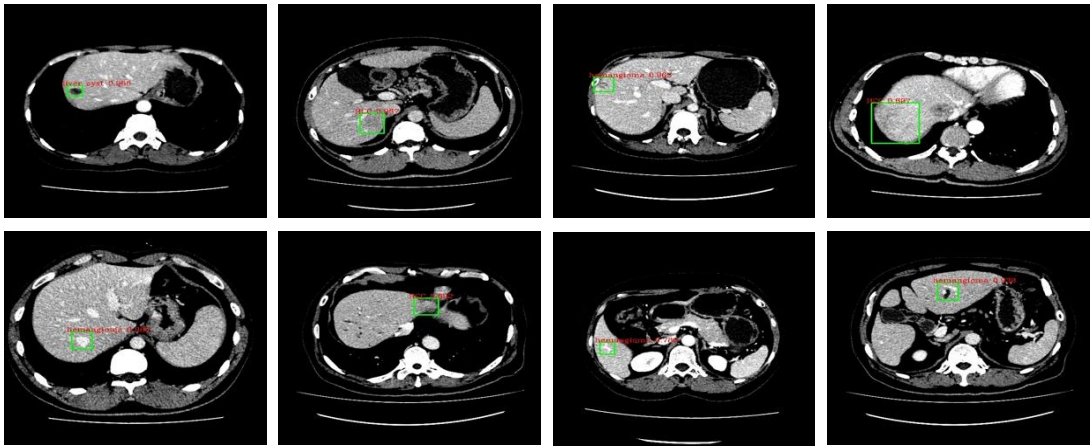

Figure 5: Examples of liver lesion-detection results. The first output row output is a good result; the second output row is a false-positive result.

**Conclusions**

The detection of features of medical images is a challenging new discipline that is the intersection of the medical field and the computing field. The work centers on this topic, using the recent deep neural network framework. Different numbers of layers are used in the neural network to extract the features of medical images to improve the accuracy of medical image detection. In the feature-extraction stage, 2D feature maps are combined with multiple slices to form a 2.5D-like structure to solve the problem of spatial information loss between different slices. During the region-proposal stage, a multi-scale-based deep network approach is used, which makes it possible to combine high-dimensional and low-dimensional information to become more sensitive to small objects. Finally, an unprecedented innovative algorithm is proposed that uses neural networks to learn the relationships between multiple modalities and adds the relationship to the framework. In summary, we have found a model framework that can better handle automated lesion detection, which also have the vital clinical significance in early detection of hepatocellular cancer.

**Data Availability**

Raw data are available from the corresponding author upon request.

**Conflicts of Interest**

All authors declare that they have no conflicts of interest regarding the publication of this paper.

**Funding Statement**

This study was not funded.